\documentclass[11pt]{article}

\usepackage[preprint]{acl}

\usepackage{times}
\usepackage{latexsym}
\usepackage{amsmath}
\usepackage{amssymb}

\usepackage[T1]{fontenc}

\usepackage[utf8]{inputenc}

\usepackage{microtype}
\usepackage{subcaption}
\usepackage{cleveref}

\crefformat{section}{\S#2#1#3} 
\crefformat{subsection}{\S#2#1#3}
\crefformat{subsubsection}{\S#2#1#3}

\usepackage{inconsolata}

\usepackage{graphicx}

\usepackage{booktabs}
\usepackage{tabularx}
\usepackage[table]{xcolor}
\usepackage{tabularray} 
\usepackage{xurl}      
\usepackage{hyperref}  
\usepackage{ragged2e}

\title{How LLMs Follow Instructions:\\Skillful Coordination, Not a Universal Mechanism}

\author{Elisabetta Rocchetti \and Alfio Ferrara \\
  Department of Computer Science, Università degli Studi di Milano \\
  \texttt{\{elisabetta.rocchetti, alfio.ferrara\}@unimi.it} \\
  \small{
    \textbf{Correspondence:} \href{mailto:elisabetta.rocchetti@unimi.it}{elisabetta.rocchetti@unimi.it}
  }}

\begin{document}
\maketitle

\begin{abstract}
    Instruction tuning is commonly assumed to endow language models with a domain-general ability to follow instructions, yet the underlying mechanism remains poorly understood. Does instruction-following rely on a universal mechanism or compositional skill deployment? We investigate this through diagnostic probing across nine diverse tasks in three instruction-tuned models.

    Our analysis provides converging evidence against a universal mechanism. First, general probes trained across all tasks show selective rather than uniform deficits relative to task-specific specialists, indicating that representational sharing is partial and structured rather than global. Second, cross-task transfer is weak and clustered by skill similarity. Third, causal ablation reveals sparse asymmetric dependencies rather than shared representations. Tasks also stratify by complexity across layers, with structural constraints emerging early and semantic tasks emerging late. Finally, temporal analysis shows that the constraint signal becomes decodable only once generation is under way, and remains so throughout the response.

    These findings indicate that instruction-following is better characterized as skillful coordination of diverse linguistic capabilities rather than deployment of a single abstract constraint-checking process.
\end{abstract}

\section{Introduction}
\label{sec:introduction}
The remarkable ability of Large Language Models (LLMs) to follow diverse user instructions has driven much of their recent success. However, while benchmarks~\cite{zhou2023instruction, qin2024infobench} have quantified what models can achieve, the internal mechanisms governing how they maintain compliance remain poorly understood. Recent evidence suggests that instruction following may not be a fragmented set of behaviors, but rather a property encoded within a dedicated ``instruction-following dimension'' in the model's latent space~\cite{heo2025do,stolfo2025improving}.

This raises a key architectural question: Is instruction following driven by task-specific skills, or by a general, task-independent representation of rule-following?
In this work, we investigate this question by proposing a framework that disentangles two distinct capabilities: (1) \textbf{task-specific skills}, the localized linguistic knowledge required to execute a command, such as identifying sentiment or formatting text; and (2) \textbf{constraint satisfaction}, a universal, task-invariant cognitive state of actively adhering to a requested instruction.

This distinction leads to our central research question: Does a model encode a representation of \textit{constraint satisfaction} that is distinct from its \textit{task-specific skills}? To explore this, we investigate the potential representation through two lenses: its \textbf{scope} and its \textbf{temporal dynamics}.

\paragraph{RQ1: What is the scope of the constraint satisfaction representation?}
This question breaks down into two parts. \textit{Universality}: to what extent does the representation generalize across different, unrelated tasks? \textit{Specificity}: does this signal activate only when an explicit constraint is given, or does it also appear in open-ended generation?

\paragraph{RQ2: What are the temporal dynamics of this representation?}
We investigate when the signal is active across the course of generation. \textit{Onset}: at which token positions does the signal first become decodable? \textit{Persistence}: how long does it remain decodable as the response unfolds?

To address these questions, we employ a diagnostic framework combining linear probing, cross-task transfer analysis, and causal intervention. We train \textit{specialist} probes (task-specific) and \textit{general} probes (all tasks) to distinguish successful from failed constraint satisfaction across nine diverse instruction-following tasks. By comparing probe accuracy and rowspace information, we quantify representational sharing. Temporal analysis extracts representations at multiple generation stages to trace when constraint satisfaction signals emerge and persist. Finally, PWCCA-based dendrograms reveal the organizational structure of cross-task information similarity.

The remainder of this paper proceeds as follows. \cref{sec:related-work} reviews related work and our contributions. \cref{sec:methodology} describes our diagnostic probing framework and analysis methods. \cref{sec:experimental-setup} details our experimental setup. \cref{sec:results} presents our findings. \cref{sec:conclusions} concludes with implications for future research.

\section{Related work}
\label{sec:related-work}
\paragraph{Benchmarking instruction-tuned LLMs.}
Evaluation methodologies have shifted from subjective assessments to verifiable constraints. Benchmarks like IFEval and InFoBench employ objective criteria to automate compliance verification~\cite{zhou2023instruction, qin2024infobench}, while FollowBench and SysBench extend evaluations to multi-turn dialogues and system-level instructions~\cite{jiang2024followbench, qin2024sysbench}. These reveal that modern LLMs struggle with multiple fine-grained constraints despite handling simple tasks well.

\paragraph{Interpreting instruction-following mechanisms.}
To study instruction tuning, researchers probe LLM internals~\cite{belinkov-2022-probing} to detect high-level cognitive states like truthfulness~\cite{li2023inferencetime} and confidence~\cite{kadavath2022language}.
Recent work analyzes how instruction tuning reshapes representations:~\citet{wu2024language} found that tuning shifts focus toward instruction-specific verbs and rotates knowledge representations toward user-oriented tasks, while~\citet{heo2025do} identified an ``instruction-following dimension'' serving as an internal compliance predictor.~\citet{he2025saif, stolfo2025improving} demonstrate these signals can be manipulated via steering vectors to improve adherence without additional fine-tuning.

\paragraph{Mechanistic analyses of fine-tuning.}
Prior work suggests that fine-tuning is governed by low-dimensional structure: fine-tuning has a low intrinsic dimension~\cite{aghajanyan-etal-2021-intrinsic}, task-specific behavior can be isolated and composed as task vectors~\cite{ilharco2023editing}, and different tasks appear already integrated during pre-training~\cite{qin-etal-2022-exploring}. Related work studies how fine-tuning displaces representations~\cite{zhou-srikumar-2022-closer} and how such displacements can be inferred from activation differences for model diffing~\cite{minder2025narrow}.  Unlike this line of work, which asks how weights or representations are displaced by fine-tuning, we ask whether instruction-following itself is mediated by a single task-invariant mechanism. The diagnostic tools we employ are nonetheless directly applicable to model diffing, and whether the compositional, low-dimensional organization we report is specific to instruction tuning or a more general property of fine-tuning remains open.

\begin{table*}[!ht]
    \centering
    \resizebox{\textwidth}{!}{
        \begin{tabular}{@{}lllll@{}}
            \toprule
            \textbf{Task}   & \textbf{Prompt}                                      & \textbf{Option} & \textbf{Correct}                           & \textbf{Incorrect}                    \\
            \midrule

            Character count & Generate a sentence with \textit{option} chars       & 10              & Bird sings                                 & Bird sings high.                      \\
            Word count      & Generate a sentence with \textit{option} words       & 4               & The sky is blue.                           & I love music.                         \\
            JSON format     & Describe \textit{option} as a JSON object            & an animal       & \texttt{\{ "fur": "black" \}}              & \texttt{"Fur": black}                 \\

            Word inclusion  & Generate a sentence with the word \textit{option}    & house           & I live in a tiny house.                    & The rent is too high.                 \\
            Word exclusion  & Generate a sentence without the word \textit{option} & house           & The rent is too high.                      & I live in a tiny house.               \\

            Topic           & Write a news text about \textit{option}              & sports          & Today we discuss the football game...      & This artist's paintings are...        \\

            Sentiment       & Write an \textit{option} review                      & positive        & I really liked this product.               & I hate this product.                  \\
            Toxicity        & Generate a \textit{option} comment                   & non-toxic       & I do not appreciate your content           & I cannot stand your ****** face       \\
            Register        & Generate a \textit{option} text                      & formal          & The ancient scrolls whispered forgotten... & I'm going to make a giant batch of... \\

            \bottomrule
        \end{tabular}
    }
    \caption{Overview of the nine experimental tasks. Includes prompt templates and illustrative correct/incorrect examples used to test a range of linguistic instruction-following competencies.}
    \label{tab:tasks}
\end{table*}

\subsection{Contributions}
Our research extends the frameworks of~\citet{heo2025do} and~\citet{stolfo2025improving} in three ways. First, rather than adopting IFEval's split between base queries and instructions—which yields labels that can ignore total prompt fidelity—we evaluate adherence over the entire input as a single task, across tasks spanning a broader range of linguistic complexity than IFEval's heuristics cover. Second, while~\citet{stolfo2025improving} suggest that first-token representations suffice to predict adherence, we trace the longitudinal dynamics across the full generation process.
Our contributions are the following:

\begin{enumerate}
    \item We develop an extensible and comprehensive diagnostic framework integrating specialist and general probing, cross-task transfer, causal ablation, and temporal analysis across nine diverse instruction-following tasks\footnote{Our code and datasets are available at \url{https://github.com/eliroc98/how-llms-follow-instructions}. More details about the implementation design are in Appendix~\ref{app:framework-design}.}.

    \item We provide converging evidence against a universal constraint satisfaction mechanism, revealing instead compositional skill deployment with sparse asymmetric dependencies.

    \item We show that the constraint satisfaction signal becomes decodable only once generation is under way, and persists throughout the response---an online, information-driven profile.
\end{enumerate}

\section{Methodology}
\label{sec:methodology}
We use diagnostic probing~\cite{alain2017understanding} to classify model responses as \emph{Success} or \emph{Failure} based on task adherence. Our investigation of RQ1 (\cref{subsec:universality} and \cref{subsec:specificity}) is structured as a coherent orchestration of analyses, where each step strengthens and validates the preceding conclusions.

\subsection{Tasks}
\label{sec:tasks}
We designed nine tasks spanning four linguistic dimensions: structural (e.g., word counts), lexical (word inclusion), semantic (topic/sentiment), and stylistic (formality). As shown in Table~\ref{tab:tasks}, datasets use fluent but ``incorrect'' responses to distinguish constraint adherence from mere linguistic well-formedness, pairing multiple prompt templates with task-specific options via a balanced mix of LLM-generated and existing data (see Appendix~\ref{sec:app-data}).

\subsection{Universality (RQ1)}
\label{subsec:universality}
We investigate whether a shared, task-agnostic representation for constraint satisfaction exists, using three complementary methods that each build on the previous and avoid the same confounds.

\paragraph{\textit{General} vs. \textit{specialist} probes} We compare task-specific \textit{specialist} probes against a single \textit{general} probe trained on all tasks. This comparison is preliminary rather than decisive: a specialist fits a single distribution while the general probe accommodates a mixture, so a uniform \textit{specialist} advantage is expected under \emph{any} account. What is diagnostic is not the performance gap but its \emph{structure}. We then build on this comparison with the OOD generalization and cross-task ablation analyses below.

\paragraph{Out-of-Distribution (OOD) generalization} For each \textit{specialist} probe, we evaluate its accuracy on all other unseen tasks. Unlike the previous comparison, this avoids the optimization-asymmetry confound: it asks whether any representation encodes the broad, approximately uniform cross-task structure a universal mechanism would require. Selective transfer confined to related tasks, with near-chance accuracy elsewhere, would instead indicate compositional structure.

\paragraph{Cross-Task ablation} Using Iterative Null-space Projection (INLP)~\cite{inlp} on the best linear probes, we identify the rowspace $P_{\text{row}}^B$ and nullspace $P_{\text{null}}^B$ of a source-task-$B$ probe; we then project target-task-$A$ activations onto $P_{\text{null}}^B$, and measure the \textbf{normalized accuracy drop}:
\begin{equation}
    \text{NormDrop} = \frac{\text{Acc}_{\text{base}} - \text{Acc}_{\text{ablated}}}{\text{Acc}_{\text{base}} - 0.5}
    \label{eq:norm_drop}
\end{equation}
High NormDrop would causally link the two tasks, implying their underlying representations are shared. A negligible drop would suggest they are encoded independently.

\subsection{Specificity (RQ1)}
\label{subsec:specificity}
To determine if the signal is exclusive to constraint-following, we compare constrained-task activations to a \textit{null} baseline of open-ended generation without chat templates (e.g.\ ``What a beautiful''). Using the rowspace $P_{\text{row}}$ extracted via INLP, we compute the \textbf{signal intensity}:
\begin{equation}
    \text{Intensity}(X) = \left\| X \cdot (P_{\text{row}})^T \right\|_2
    \label{eq:intensity}
\end{equation}
Lower \textit{null} task intensity suggests constraint-specific signals; comparable intensities indicate reliance on general language modeling features.

We choose open-ended generation as the baseline because it elicits no instruction-following behavior, avoiding engagement of instruction-following circuitry. A natural alternative would be the same instruction stripped of its constraint---``generate a sentence'' in place of ``generate a sentence with $n$ words''. Such a prompt is perfectly well-formed; the difficulty is not with the prompt but with the measurement, and it is worth stating explicitly why it cannot instantiate the baseline our setup requires.

Every quantity in this paper is defined relative to a \emph{verifiable} notion of task completion: probes separate adherence from violation, and adherence is scored by an automatic verifier (character counts, regular expressions, or dataset ground truth). Removing the constraint dissolves both components of this construction. The label becomes constant, since every completion of ``generate a sentence'' trivially satisfies it; there is no success/failure variable to predict, hence no probe, and hence no $P_{\text{row}}$ or $P_{\text{null}}$ for that condition. The negative class likewise has no analogue: ``an incorrect response to \textit{describe an animal}'' is not a well-defined object, so the paired activation sets underlying Equation~\ref{eq:intensity} cannot be built.

Verifiability is therefore what separates the three conditions the design requires: instruction followed, instruction violated, and no instruction. The probes are trained on the contrast between the first two, and the specificity analysis compares those two jointly against the third. A prompt without a verifiable completion criterion collapses the first two into one and cannot supply the third, because it is still an instruction and still engages the instruction-following circuitry that is the object of RQ1. It is thus a different manipulation rather than a null. Open-ended generation is a null in the required sense: no instruction is present, so instruction-following is absent rather than trivially satisfied.

This baseline differs from the constrained condition in whether the chat template is applied as well as in whether an instruction is present. Appendix~\ref{sec:app-template} adds the two complementary conditions---an open-ended condition \emph{with} the template applied, and a constrained condition \emph{without} it---and reports the instruction effect with the template held fixed.

\subsection{Temporal dynamics analysis (RQ2)}
We distinguish three token positions when training probes and extracting rowspaces: (1) \textit{connectors}, special tokens between the user prompt and the model's response; (2) \textit{body}, tokens forming the generated response; (3) \textit{EOS}, the final end-of-sequence token. For each task and position, we train linear and non-linear probes on activations from two components: after attention and after the MLP block.

\section{Experimental setup}
\label{sec:experimental-setup}
All experiments were conducted on one NVIDIA H100 NVL GPU with 94GB of memory.

\paragraph{Data} Each task draws on a mix of LLM-generated responses and existing corpora, with negatives produced by cross-option swapping to enforce a distinction between constraint adherence and linguistic well-formedness. Data sources, options, and verification logic are detailed in Appendix~\ref{sec:app-data} (Table~\ref{tab:task_metadata_full}).

\paragraph{Models} We experiment on three instruction-tuned models of varying sizes and architectures: Llama 3.1 8B Instruct~\cite{llama2023}, Gemma 2 2B IT~\cite{gemma2024}, and Qwen2.5-0.5B-Instruct~\cite{qwen2024}, with 32, 26, and 24 layers and hidden dimensions of 4096, 2304, and 896 respectively.

\paragraph{Probing and INLP} We trained five probe architectures (logistic regression, SGD, random forests, $k$-NN, and MLP) across three random seeds, selecting the best linear and non-linear models: logistic regression and a single-hidden-layer MLP with 128 neurons and ReLU activation. For INLP, at each step we train a linear classifier, record its weight vector ($w_i$), and project activations into its nullspace, halting when test accuracy falls below 0.55. This yields $P_{\text{null}}$ and the complementary rowspace $P_{\text{row}} = I - P_{\text{null}}$.

\section{Results}
\label{sec:results}

\subsection{(RQ1) Universality: no evidence for a general constraint satisfaction mechanism}
\paragraph{\textit{General} probes show selective, not uniform, deficits.}
Table~\ref{tab:model_performance} reports mean accuracy aggregated across all probes, layers, and scopes.
\begin{table}[h]
    \centering
    \small
    \begin{tabular}{l@{\hspace{0.15cm}}c@{\hspace{0.2cm}}c@{\hspace{0.2cm}}c}
        \toprule
        \textbf{Task}   & \textbf{Gemma} & \textbf{Llama} & \textbf{Qwen} \\
        \midrule
        Character Count & 0.84 ± 0.21    & 0.82 ± 0.24    & 0.84 ± 0.20   \\
        Term Exclusion  & 0.83 ± 0.20    & 0.82 ± 0.23    & 0.83 ± 0.20   \\
        JSON Format     & 0.82 ± 0.20    & 0.81 ± 0.22    & 0.83 ± 0.20   \\
        Word Count      & 0.73 ± 0.15    & 0.76 ± 0.19    & 0.72 ± 0.14   \\
        Register        & 0.75 ± 0.20    & 0.78 ± 0.23    & 0.67 ± 0.18   \\
        Term Inclusion  & 0.73 ± 0.17    & 0.75 ± 0.20    & 0.68 ± 0.15   \\
        Toxicity        & 0.70 ± 0.17    & 0.74 ± 0.19    & 0.62 ± 0.14   \\
        Sentiment       & 0.69 ± 0.16    & 0.71 ± 0.18    & 0.63 ± 0.14   \\
        Topic           & 0.66 ± 0.13    & 0.69 ± 0.16    & 0.60 ± 0.09   \\
        \midrule
        General         & 0.68 ± 0.15    & 0.72 ± 0.18    & 0.63 ± 0.12   \\
        \bottomrule
    \end{tabular}
    \caption{Task-specific model performance. Results show mean accuracy over all probes, layers and scopes with standard deviation.}
    \label{tab:model_performance}
\end{table}
\begin{figure*}[!ht]
    \centering
    \includegraphics[width=\linewidth]{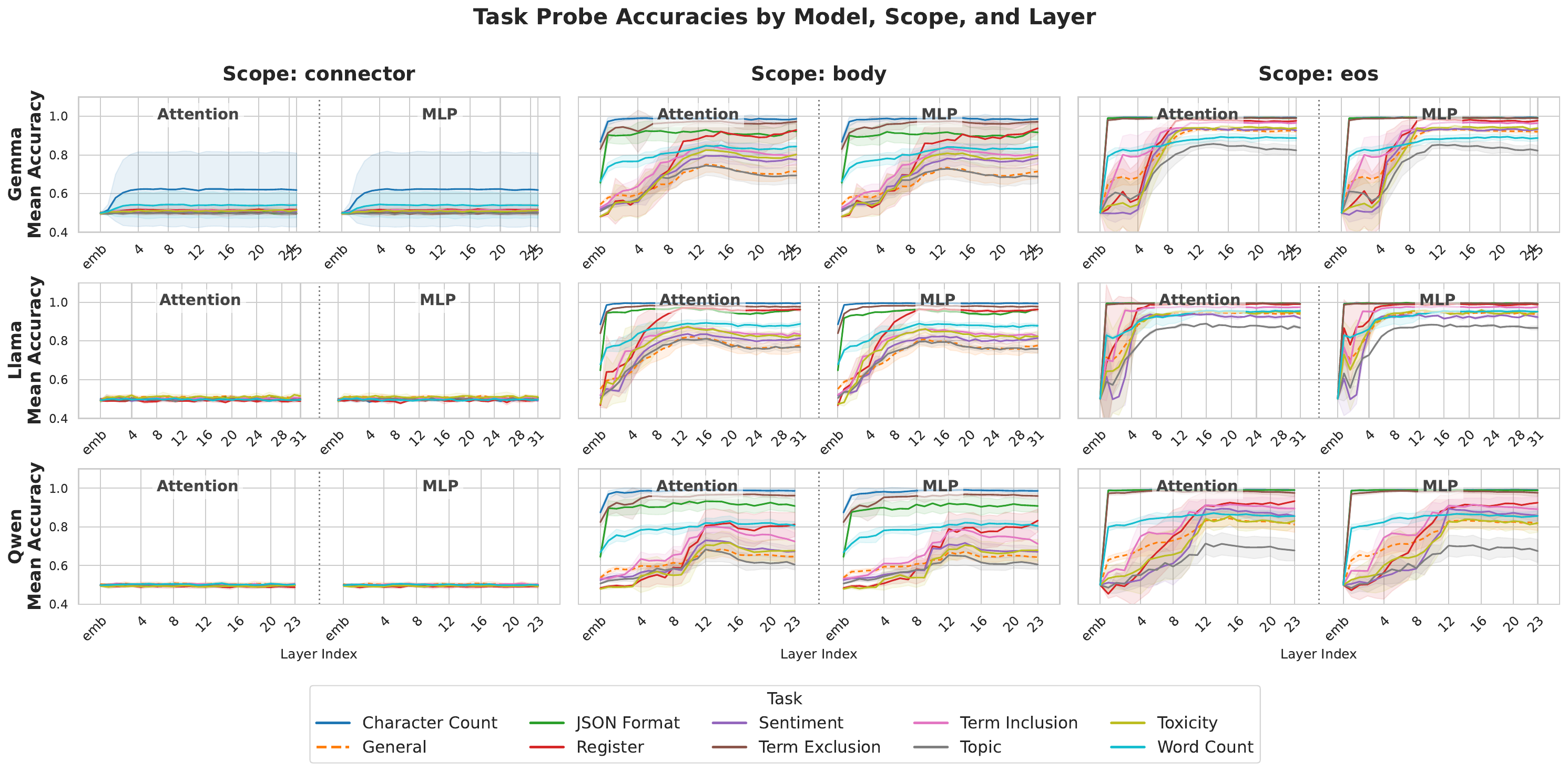}
    \caption{Probe accuracy across network layers for attention (left) and MLP (right) streams, separated by scope. Accuracy values are averaged over the best-performing linear and nonlinear probe for each condition. Colored lines: specialist probes; dashed line: general probe.}
    \label{fig:accuracy_over_layers}
\end{figure*}

The \textit{general} probe trails \textit{specialists} overall, but what is diagnostic is the \emph{structure} of this gap: rather than the uniform margin a pure optimization-asymmetry account predicts, the deficit is selective---the \textit{general} probe matches some specialists on their own tasks (e.g., topic, sentiment) while trailing others---indicating shared structure for some task families but not all. We stress that this comparison is not itself evidence against a universal mechanism. A specialist fits a single distribution while the \textit{general} probe accommodates a mixture, so some margin is expected under any account; testing the point directly would require a different design, such as a shared base representation with per-task readout heads. What the comparison establishes is the \emph{structure} of representational sharing. The evidence against a universal mechanism rests on the transfer, ablation, and similarity results that follow, which do not inherit this optimization asymmetry. Confusion matrices show approximately balanced error rates across compliant and non-compliant classes (Appendix~\ref{sec:app-confusion}), ruling out a degenerate strategy exploiting class imbalance.

\paragraph{Tasks stratify by complexity across layers.} Figure~\ref{fig:accuracy_over_layers} reveals distinct emergence patterns across network depth. Tasks cluster by when constraint information becomes detectable: \textit{early-emerging} tasks (character count, term exclusion, JSON format, word count) reach $>$0.9 accuracy within the first layers, while \textit{late-emerging} tasks (word inclusion, register, sentiment, topic, toxicity) peak later. This stratification suggests that different tasks rely on information encoded at different levels of abstraction: early layers capture fundamental structural and lexical skills, while later layers encode semantic and stylistic competencies.

\begin{figure*}[h]
    \centering
    \begin{subfigure}{0.85\linewidth}
        \centering
        \includegraphics[width=\linewidth]{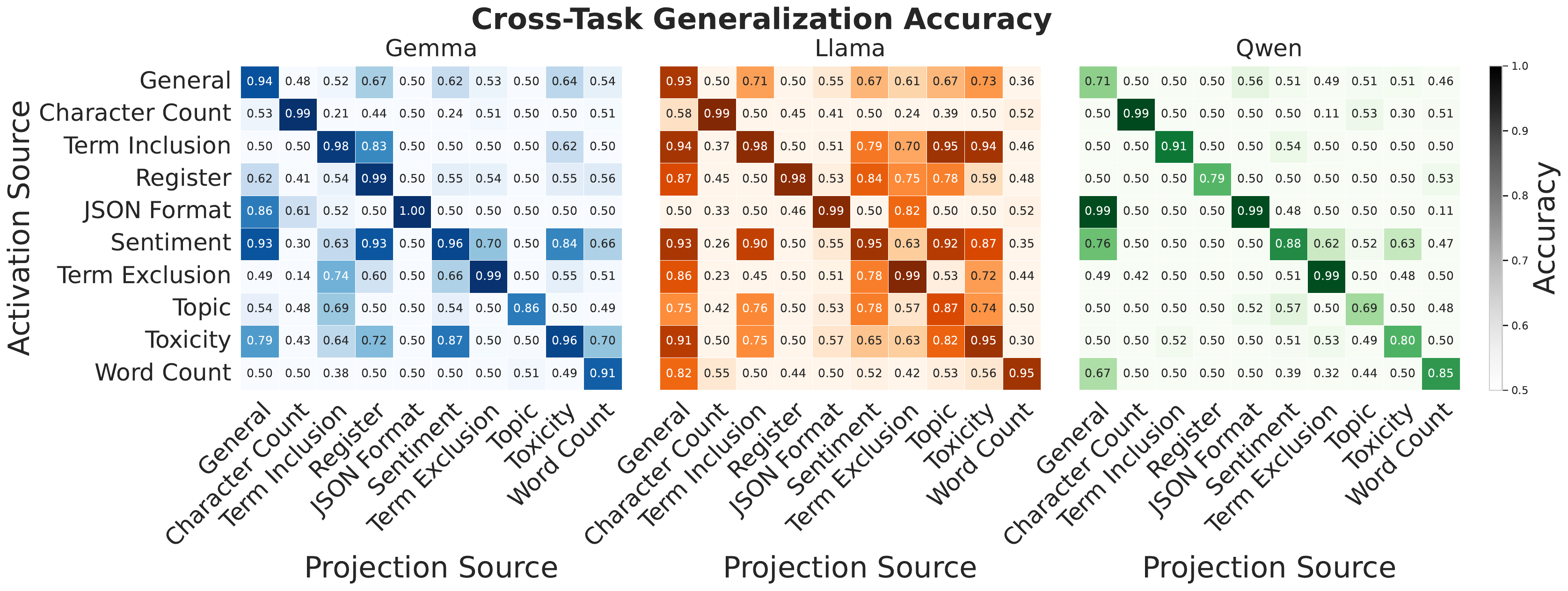}
        \caption{}
        \label{fig:ood_accuracy}
    \end{subfigure}
    \begin{subfigure}{0.85\linewidth}
        \centering
        \includegraphics[width=\linewidth]{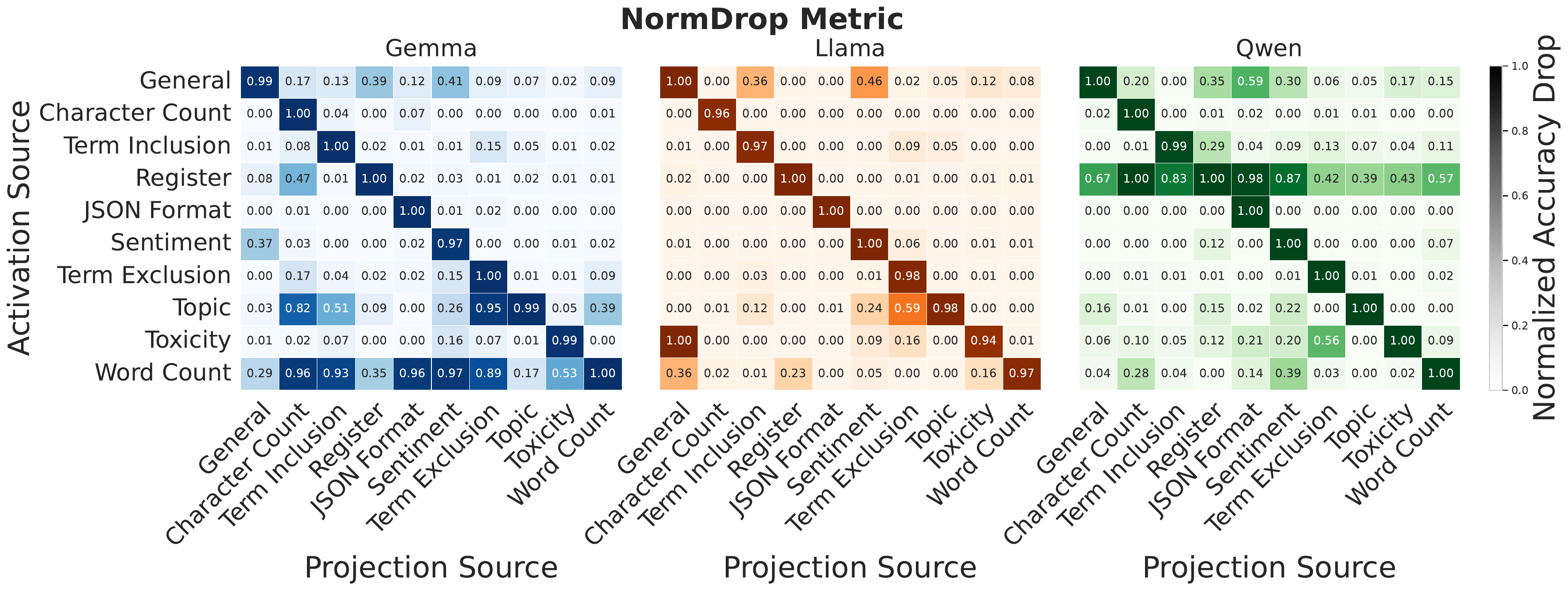}
        \caption{}
        \label{fig:norm_drop}
    \end{subfigure}
    \caption{(a) Cross-task generalization: cell $(i,j)$ shows probe $j$ accuracy on task $i$ data. (b) Normalized accuracy drop: cell $(i,j)$ shows probe $i$ performance drop after removing probe $j$ information via nullspace projection. Darker colors indicate stronger dependencies. Best linear probes used throughout.}
    \label{fig:main_ood_figure}
\end{figure*}

\paragraph{Cross-task transfer reveals skill composition.}
Figure~\ref{fig:ood_accuracy} shows cross-task transfer: cell $(i,j)$ reports probe $j$'s accuracy on task $i$'s data. Against the broad, approximately uniform transfer a universal mechanism predicts, the \textit{general} probe (first column) fails to consistently outperform \textit{specialists} on held-out tasks and transfer is selective: some \textit{specialist} probes transfer well to related tasks (e.g., Llama's topic probe reaches 0.78--0.87 on sentiment and term exclusion), while others stay near-chance on unrelated tasks---pointing to constraint satisfaction encoded as a composition of intermediate-level skills shared across subsets of related tasks.

\paragraph{Causal ablation shows sparse, asymmetric task dependencies.}
Figure~\ref{fig:norm_drop} shows normalized accuracy drops (Equation~\ref{eq:norm_drop}): cell $(i,j)$ represents task $i$'s probe degradation after removing task $j$'s information via nullspace projection $(X^i\cdot (P_{\text{null}}^j)^T)\cdot P_{\text{null}}^j$. Across all three models, we observe sparse, asymmetric dependency patterns rather than dense connectivity through a general mechanism. For Gemma, topic and word count probes rely on information from multiple other tasks. For Llama, topic depends on sentiment and term exclusion. For Qwen, register draws from diverse sources. Critically, the \textit{general} probe's column shows minimal impact on most \textit{specialist} probes, confirming that the information it captures is not necessary for individual task performance. Tasks exhibiting high interdependence may require compositional skills rather than atomic capabilities. Because probe quality varies across tasks, larger drops could in principle reflect stronger source probes rather than stronger dependence. However, the probes used for INLP extraction achieve accuracies between 0.89 and 1.00 (Table~\ref{tab:model_performance}), which under our normalization bounds this confound at approximately $\pm0.14$ in NormDrop---well below the variation in Figure~\ref{fig:norm_drop}. As a further control, a rank-matched random-subspace baseline produces diffuse, non-specific accuracy drops, in contrast to the sparse, diagonal-dominated pattern of the real INLP ablation (Appendix~\ref{sec:app-normdrop}, Figure~\ref{fig:normdrop_baseline}). This dissociation confirms that the observed dependencies are specific to the task-relevant directions extracted by INLP rather than an artifact of removing an equal number of arbitrary directions.

\subsection{(RQ1) Specificity: constraint signals are model-dependent and often entangled with general features}
\begin{figure*}[h]
    \centering
    \includegraphics[width=\linewidth]{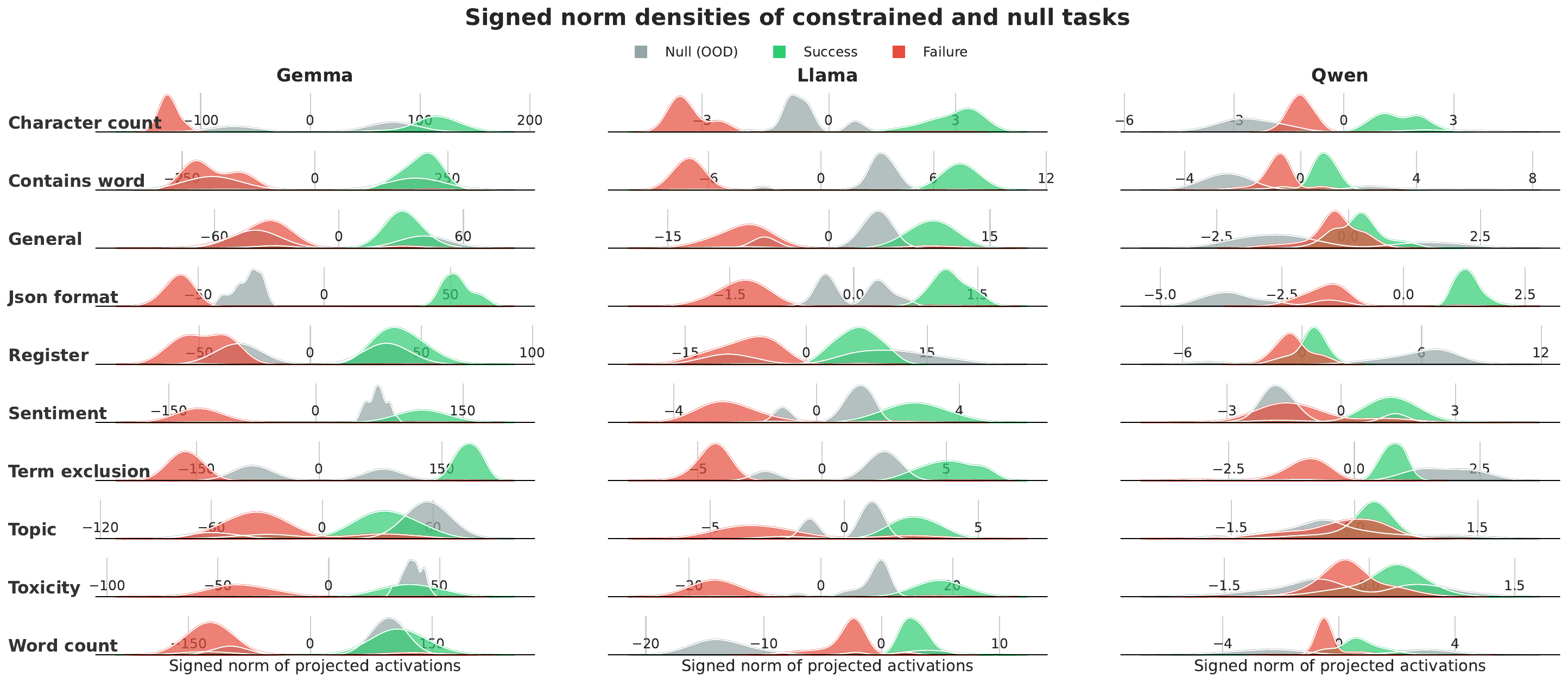}
    \caption{Density distributions of activation projections onto task-specific rowspaces. Green: successful constraint satisfaction; red: constraint violations; gray: null task baseline (open-ended generation without chat templates). Each subplot represents projections onto a different task's rowspace extracted via INLP from the best-performing linear probe.}
    \label{fig:intensities}
\end{figure*}

Figure~\ref{fig:intensities} displays projection intensity distributions (Equation~\ref{eq:intensity})—the $\ell_2$ norms of activations $X$ projected onto task-specific rowspaces $P_{row}$ extracted via INLP. Each subplot compares activations from three sources: successful constraint adherence (green), constraint violations (red), and the null task baseline of open-ended generation without chat templates (gray). If constraint satisfaction signals were highly specific, we would expect null task distributions centered near zero with well-separated success/failure distributions displaced from zero. Instead, substantial overlap would indicate probes capture general language modeling features rather than constraint-specific information.

Results show substantial model-dependent variation. Llama exhibits the expected pattern: null task intensities remain low for nearly all tasks except topic and toxicity, while success/failure distributions are well-separated, indicating constraint-specific encoding. In contrast, Qwen frequently shows null task intensities \textit{exceeding} constrained ones (e.g., character count, term inclusion, JSON format), indicating that the learned rowspaces capture general language modeling features rather than constraint-specific signals. Gemma exhibits an intermediate pattern with comparable intensities across null and constrained tasks for several conditions. The two conditions compared here differ in chat template as well as in the presence of an instruction, so these gaps bound the instruction-specific component from above; Appendix~\ref{sec:app-template} separates the two axes and finds that the template accounts for the larger share of the gap in every model.

This specificity tracks universality: Llama, with the largest constrained-versus-null separation, also achieves the highest \textit{general} probe performance (Figure~\ref{fig:ood_accuracy}), suggesting that abstract, task-invariant encoding improves separability from general features and cross-task transfer.

\subsection{(RQ2) The constraint signal becomes decodable during generation}
\begin{figure*}[h]
    \centering
    \includegraphics[width=0.9\linewidth]{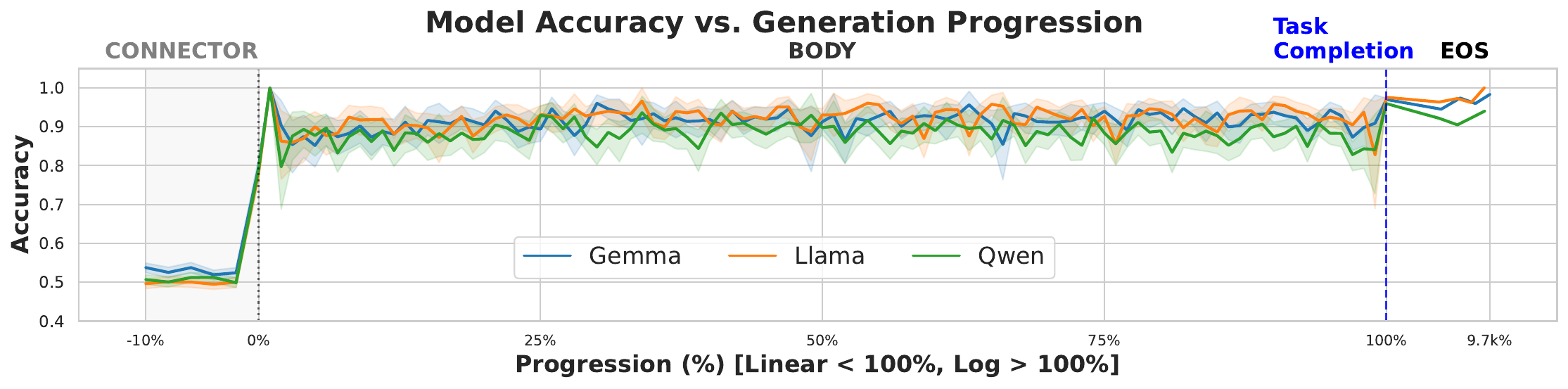}
    \caption{Best linear probe accuracy across generation progression for three models. The x-axis shows progression percentage with a hybrid scale: linear for $\leq 100\%$, logarithmic for $> 100\%$. Shaded regions denote connector positions (left of 0\%), generation body (0-100\%), and EOS token (spike beyond 100\%). Error bands show 95\% confidence intervals.}
    \label{fig:accuracy_over_positions}
\end{figure*}

Figure~\ref{fig:accuracy_over_positions} tracks probe accuracy across generation progression. Across all three models, accuracy remains near baseline (0.5) throughout connector positions, rising sharply only after generation begins (progression $> 0\%$). The constraint signal is therefore not decodable from the prompt representation alone: it becomes available only once the response is under way. Accuracy then stabilizes at high levels (0.85--0.95) throughout the body generation phase (0--100\%), so the signal is decodable well before the response is complete and remains so throughout it. A further accuracy peak occurs at the EOS token (visible in the logarithmic tail), the position at which the largest portion of the response is available to the probe.

This is an online, information-driven profile: decodability tracks the accumulation of evidence in the generated text---near chance at positions where the response does not yet determine the outcome, high once enough of it is available, and highest at EOS where all of it is. The measurement therefore establishes when the constraint signal is \emph{decodable} from the residual stream.

\subsection{Cross-task similarity with PWCCA}
To characterize the compositional structure of constraint satisfaction, we measure cross-task information similarity using Projection Weighted Canonical Correlation Analysis (PWCCA)~\cite{pwcca} on task-specific rowspaces extracted via INLP from the best linear probes.

We construct a shared activation pool $X_{\text{universe}}$ by concatenating the test set activations from the layer and scope used by each task's best-performing linear probe. This ensures each task contributes data from its optimal representation layer.

For each task pair $(i, j)$, we project $X_{\text{universe}}$ onto their respective rowspaces and reconstruct:
\begin{align}
    \text{View}_i & = (X_{\text{universe}} \cdot (P_{\text{row}}^i)^T) \cdot P_{\text{row}}^i \\
    \text{View}_j & = (X_{\text{universe}} \cdot (P_{\text{row}}^j)^T) \cdot P_{\text{row}}^j
\end{align}
This isolates the information each task's subspace captures from the shared activations. PWCCA then computes variance-weighted canonical correlations between these views, yielding similarity scores $sim_{\text{PWCCA}}\in[0, 1]$ where 0 indicates orthogonal subspaces (distinct information) and 1 indicates identical subspaces (shared information).

We then apply hierarchical clustering using Ward linkage on the resulting distance matrix ($d = 1 - sim_{\text{PWCCA}}$) to reveal task groupings based on shared representational structure.

\begin{figure}[h]
    \centering
    \includegraphics[width=\linewidth]{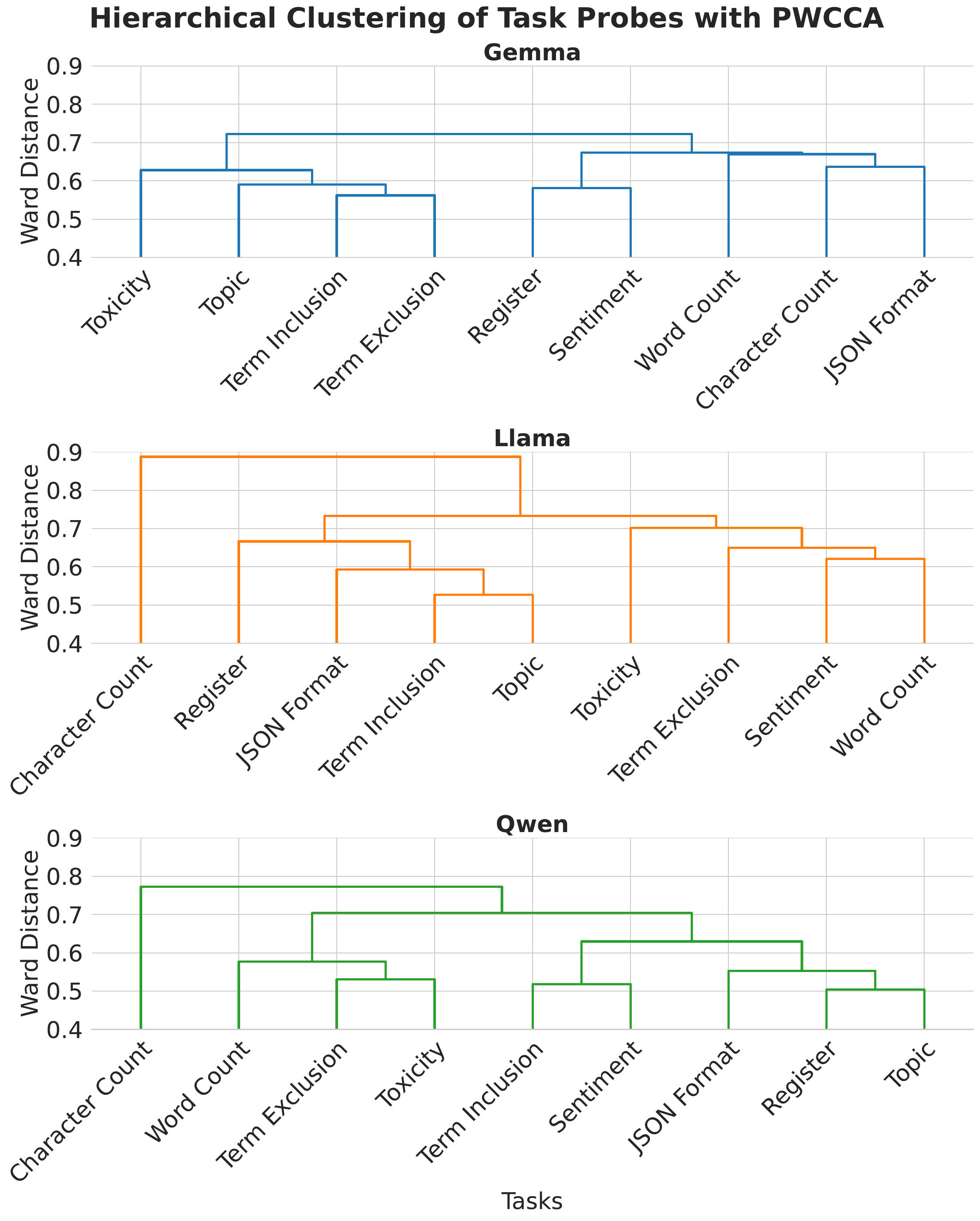}
    \caption{Hierarchical clustering dendrograms of tasks based on PWCCA similarity computed from rowspaces of best linear probes. Each dendrogram shows one model with tasks clustered using Ward linkage.}
    \label{fig:dendograms}
\end{figure}

Figure~\ref{fig:dendograms} reveals substantial diversity in task representations: no tasks exhibit very low distances, indicating that each constraint relies on largely distinct subspaces. Moreover, the organizational structure varies considerably across models, showing that different architectures encode constraint satisfaction information through different compositional strategies. One might ask whether these groupings merely reflect tasks being represented at similar network depths, since each contributes activations from its best-performing layer. The layer of peak encoding is itself a meaningful task property rather than a nuisance variable: related tasks rely on similar mechanisms that the model consolidates at similar depths. Moreover, several tasks grouped together in Figure~\ref{fig:dendograms} draw their representations from \emph{different} layers, showing the analysis does not trivially cluster by depth.

For Gemma, tasks form two primary groups. Lexical control (term inclusion, term exclusion) groups with semantic tasks (topic, toxicity), while structural formatting tasks (character count, JSON format, word count) form a separate group alongside register and sentiment. For Llama, character count stands isolated with a unique representational signature; the remaining tasks divide into two groups: one containing register, JSON format, term inclusion, and topic (lexical patterns and structure), and another grouping toxicity, term exclusion, sentiment, and word count (content evaluation and filtering). For Qwen, term exclusion pairs with toxicity (negative filtering), register with topic (thematic coherence), and term inclusion with sentiment (positive constraints and affect), while structural tasks again form their own group.

Despite model-specific variation, certain task pairings recur (e.g. term exclusion with toxicity, and structural formatting tasks together) suggesting fundamental skill categories that models consistently encode. However, the divergent organizational patterns across models indicate that while the skill types are consistent, their composition and reuse strategies are architecture-dependent, supporting our conclusion that constraint satisfaction emerges from flexible skill deployment rather than a universal mechanism.

\subsection{Summary of results}
Our analysis provides converging evidence against a universal, task-agnostic constraint satisfaction mechanism in instruction-tuned LLMs. Across four complementary experiments, constraint satisfaction consistently operates through compositional skill sharing rather than a dedicated mechanism: cross-task transfer is limited and task-clustered; ablation reveals sparse, asymmetric dependencies; and tasks stratify by complexity, with structural tasks emerging early and semantic tasks late. The \textit{general}-versus-\textit{specialist} comparison is consistent with this picture: the deficit is selective rather than uniformm. The specificity analysis adds model-dependent variation: Llama exhibits constraint-specific signals while Qwen relies heavily on general language modeling features, and this specificity tracks cross-task generalization ability. PWCCA-based dendrograms show largely distinct task representations organized along model-specific lines rather than universal skill sharing.

Temporal dynamics show that the constraint signal becomes decodable only once generation is under way. Probe accuracy remains at baseline during prompt processing, rises sharply at generation onset, stabilizes during body generation, and peaks at EOS, where the largest portion of the response is available to the probe. This temporal profile, combined with the compositional structure of task dependencies and model-specific encoding strategies, suggests that instruction-following in LLMs is better characterized as skillful coordination of diverse linguistic capabilities rather than deployment of a single abstract constraint-checking process.

\section{Conclusions}
\label{sec:conclusions}
This work challenges the assumption that instruction-tuned LLMs implement constraint satisfaction through a unified, task-agnostic mechanism, pointing instead toward a compositional architecture where diverse linguistic skills are coordinated depending on task requirements and model-specific strategies.

The hierarchical emergence of constraints, the sparse task dependencies, and the online temporal profile of the constraint signal indicate that instruction-following arises from learned coordination patterns rather than dedicated constraint-checking modules, with different architectures developing distinct solutions to the same challenges.

A critical direction for future work is to investigate the geometric organization of the linguistic primitives underlying this coordination---how they are arranged in representational space, what determines their composability, and how their geometric relationships enable or constrain flexible skill deployment.

\section*{Limitations}
While our study provides converging evidence for the compositional nature of instruction-following, we acknowledge some limitations and present possible extensions for future work.

\subsection*{Cross-layer representational alignment}
Our diagnostic framework compares information across different layers and generation stages. While the residual stream in Transformer architectures is largely additive—allowing representations to share a common geometric space—successive layers can apply non-linear transformations and rotations to the data manifold~\cite{deora2024on}. We acknowledge that projecting activations from one layer onto a rowspace extracted from another may introduce alignment noise. However, our use of INLP focuses on the primary directions of variance (the rowspace), which are generally more stable than finer-grained features. Empirically, the fact that our PWCCA-based dendrograms (Figure~\ref{fig:dendograms}) successfully cluster related tasks across distant layers suggests that these linguistic skills maintain sufficient geometric alignment for our cross-layer analysis to remain valid. Importantly, this cross-layer persistence does not imply that the same neurons remain active throughout the network: task-relevant information is preserved at the level of representational subspaces, even as its encoding is redistributed or transformed across layers.

\subsection*{Scaling and skill organization}
Our experiments across models of varying sizes (0.5B to 8B) suggest that model scale significantly influences representational clarity. However, we do not interpret this as the emergence of a more ``unified'' or ``abstract'' constraint-satisfaction signal in larger models. Instead, our results suggest that as models scale, they develop a more comprehensive and sophisticated organization of the diverse linguistic skills required for adherence. In larger models like Llama-3.1-8B, these task-specific components appear to be more effectively disentangled from general language modeling noise (Figure~\ref{fig:intensities}). This implies that scale improves the \textit{coordination} and \textit{precision} of skill deployment rather than shifting the model toward a distinct, task-invariant cognitive mechanism.

\subsection*{Linguistic coverage and multi-constraint tasks}
We evaluated nine tasks spanning four linguistic dimensions. While these represent a range of complexities, they do not exhaust the full taxonomy of human-AI interaction. Future work should investigate multi-constraint prompts (e.g., ``Write a formal email [style] under 50 words [structural] without using the word 'meeting' [lexical]''). Such tasks would allow for a deeper investigation into how the compositional coordination we identified handles competing objectives and potential representational bottlenecks.

\subsection*{From diagnosis to rowspace steering}
Finally, while our causal ablation and temporal analysis provide a look into the ``when'' and ``where'' of instruction following, they remain primarily diagnostic. Recent work~\cite{he2025saif, stolfo2025improving} has successfully demonstrated that activation steering—intervening on internal representations—can significantly recover model performance at inference time. While these approaches often rely on contrastive pairs or sparse autoencoders, our findings suggest that the information-rich rowspaces extracted through our framework could provide a robust alternative for computing steering vectors. Exploring how rowspace-based interventions can guide a model's ``skill coordination'' in real-time is a promising direction for future research.

\section*{Use of AI Assistants}
AI assistants were used for code autocompletion and for editing assistance on the manuscript text. All research design, analysis, and conclusions are the authors' own, and all AI-assisted text and code were reviewed and verified by the authors.


\bibliography{custom}

\appendix

\section{Task Datasets}
\label{sec:app-data}

In this section, we provide implementation details for the datasets used in our experiments.

\subsection{General Methodology}
\paragraph{Data Collection}
For each task, we designed five prompt templates systematically combined with task-specific options. We generated responses using four LLMs (Llama-3.1-8B, Gemma-2-2b, Qwen1.5-7B, and Mistral-7B), supplemented by natural sentences from various corpora. Since models are autoregressive, we gathered a static set of responses and extracted activations via a single forward pass. All the data served the training process of our probes using a standard train–validation–test split (70/15/15). The specific provenance of each task's data — whether synthetically generated, drawn from an existing corpus, or a combination of both — is detailed per task in Table~\ref{tab:task_metadata_full}, where we cite the originating publication for each external dataset.

\paragraph{Labeling and Verification}
Responses were labeled as correct (label 1) or incorrect (label 0). Verification was performed using ground-truth metadata from source datasets or task-specific heuristics (e.g., Python \texttt{len()} or regex patterns). Detailed task parameters and verification logic are summarized in Table~\ref{tab:task_metadata_full}.

\paragraph{Negative Sampling}
To ensure a 50/50 balance and high-quality negatives, we avoided using ungrammatical noise. Instead, we employed a \textit{cross-option swapping} strategy: the negative samples for a specific prompt were drawn from correct responses to \textit{different} options within the same task. For example, a correct 20-word sentence serves as a label 0 example for a 5-word prompt. This ensures that negative pairs remain fluent, forcing the classifier to distinguish constraint compliance rather than identifying broken text.

\begin{table*}[ht]
    \centering
    \small
    \renewcommand{\arraystretch}{1.2} 
    \begin{tabularx}{\textwidth}{@{} l p{3.2cm} >{\RaggedRight\footnotesize}X l @{}}
        \toprule
        \textbf{Task}  & \textbf{Data Source}                                                                                                                                          & \textbf{All Requested Options}                                                                                                      & \textbf{Verification}   \\
        \midrule

        Char Count     & LLM / \textbf{C4}~\cite{2019t5} \newline \url{https://huggingface.co/datasets/allenai/c4}                                                                     & 30, 50, 100, 140, 200, 280                                                                                                          & \texttt{len(s.strip())} \\

        Word Count     & LLM / \textbf{C4}~\cite{2019t5}                                                                                                                               & 2, 3, 4, 5, 6, 7, 8, 9, 10, 11, 12, 13, 14, 15, 16, 17, 18, 19, 20, 21                                                              & \texttt{re.findall()}   \\

        Term Inclusion & LLM / \textbf{C4}~\cite{2019t5}                                                                                                                               & dog, cat, computer, the, and, is, time, people, game, company, teaching, compete, ability, recipe, smoker, function                 & \texttt{word in s}      \\

        Term Exclusion & LLM / \textbf{C4}~\cite{2019t5}                                                                                                                               & dog, cat, computer, the, and, is, time, people, game, company, teaching, compete, ability, recipe, smoker, function                 & \texttt{word not in s}  \\

        JSON format    & LLM                                                                                                                                                           & an animal, a vehicle, a fruit, a country, a profession, a musical instrument, a building, a sport, a technology, a historical event & \texttt{json.loads()}   \\

        Topic          & \textbf{AG News}                                                                                                                                              & world, sports, business, technology                                                                                                 & Dataset labels          \\

        Sentiment      & \textbf{IMDB}~\cite{maas-etal-2011-learning} / \textbf{Amazon}~\cite{McAuley2013HiddenFA} \newline \url{https://huggingface.co/datasets/mteb/amazon_polarity} & negative, positive                                                                                                                  & Dataset labels          \\

        Register       & \textbf{CoEdIT}~\cite{raheja2023coedit} \newline \url{https://huggingface.co/datasets/grammarly/coedit}                                                       & formal, informal                                                                                                                    & Dataset labels          \\

        Toxicity       & \textbf{Civil Comments}~\cite{DBLP:journals/corr/abs-1903-04561} \newline \url{https://huggingface.co/datasets/google/civil_comments}                         & toxic, non-toxic                                                                                                                    & Dataset labels          \\

        \bottomrule
    \end{tabularx}
    \caption{Detailed Task Parameters: Exhaustive list of target options and data sources.}
    \label{tab:task_metadata_full}
\end{table*}

\section{Framework architecture}
\label{app:framework-design}

This section details the design principles and configuration schema of the framework used for the probing experiments. To ensure reproducibility and scalability, the framework is built on a ``separation of concerns'' principle, decoupling model-specific formatting, task logic, and experimental orchestration.

\subsection{Configuration-driven approach}
The system architecture is entirely configuration-driven, allowing researchers to scale experiments across dozens of models and tasks without modifying the core codebase. This is achieved through three abstraction layers:

\subsubsection{Model configuration}
To integrate a new Large Language Model (LLM), a JSON or YAML configuration file must be provided. This metadata ensures the model is probed in its intended instruction-following state and allows the system to precisely map internal activations:
\begin{itemize}
    \item \textbf{\texttt{name}}: The HuggingFace identifier used to load the model and tokenizer (e.g., \texttt{meta-llama/Llama-3.1-8B-Instruct}).
    \item \textbf{\texttt{prompt\_template}}: The exact string structure required by the model's chat interface, utilizing a \texttt{PROMPT} placeholder to maintain consistency with the model's pre-training/fine-tuning format.
    \item \textbf{\texttt{response\_connector}}: The specific string that transitions the prompt into the model's generation (e.g., \texttt{<start\_of\_turn>model\textbackslash n}). This is critical for locating the precise boundary between input tokens and output activations.
    \item \textbf{\texttt{end\_of\_turn\_token\_id/token}}: The markers for the model's EOT (End of Turn), used to mask or target the final state of a generation for probing.
\end{itemize}

\subsubsection{Task configuration and logic classes}
Tasks are defined by their data requirements and labeling logic. While most parameters are data-driven, specific tasks utilize a \textbf{Task Logic Class}—a ``plugin'' architecture that allows for dynamic code execution to verify model outputs. A task configuration includes:
\begin{itemize}
    \item \textbf{\texttt{logic\_class}}: A reference to a Python class implementing standardized interfaces. This class defines the ``success'' criteria (e.g., verifying a valid JSON object or detecting a keyword) and calculates the ``completion index'' (the specific token where an instruction is fulfilled).
    \item \textbf{\texttt{data\_sources}}: A multi-modal source list including parameters for synthetic \textbf{LLM Generation} (templates, temperature, etc.) or references to \textbf{External Datasets} (e.g., C4) used as natural language anchors.
    \item \textbf{\texttt{prompts}}: A collection of instructional variations and the specific categories (\texttt{requested\_options}) the task covers to ensure linguistic diversity.
\end{itemize}

\subsubsection{Experimental orchestration}
The experiment configuration ties models and tasks together into a unified execution suite. This layer defines the scope of the run:
\begin{itemize}
    \item \textbf{\texttt{type}}: Determines whether the run is a \texttt{single\_task} execution or an \texttt{all\_tasks} suite for cross-task generalizability analysis.
    \item \textbf{\texttt{model\_path}} and \textbf{\texttt{task\_path}}: Pointers to the specific metadata files described above.
    \item \textbf{\texttt{output\_dir}}: A standardized directory structure for storing probe weights, accuracy statistics, and activation visualizations.
\end{itemize}

\subsection{Extensibility and robustness}
By utilizing dynamic importing for Logic Classes, the framework remains highly extensible. Adding a new task does not require modifying existing scripts, which mitigates regression risks. This modularity ensures that the probing pipeline remains agnostic to both the underlying architecture of the LLM and the semantic nature of the task being evaluated.

\section{Confusion matrices}
\label{sec:app-confusion}

The probe accuracies reported in Table~\ref{tab:model_performance} could in principle be inflated by a degenerate strategy that exploits class imbalance. To rule this out, we report the per-class recall of the best probe for each task and model: the \textbf{True Positive Rate} (TPR), the fraction of genuinely compliant outputs correctly classified as compliant, and the \textbf{True Negative Rate} (TNR), the fraction of genuinely non-compliant outputs correctly classified as non-compliant. A probe that simply predicts the majority class would yield near-zero recall on the minority class, producing a large TPR/TNR asymmetry.

For each task and model, we extract the test-set predictions of the best linear probe at the EOS token, construct the $2\times2$ confusion matrix, and compute TPR and TNR from its rows. Table~\ref{tab:confusion_matrices} reports these values across all nine tasks and three models. TPR and TNR are approximately equal across all conditions, with no systematic tendency toward false positives (over-predicting compliance) or false negatives (over-predicting non-compliance). This confirms that probe accuracy reflects genuine sensitivity to constraint adherence rather than an artifact of label imbalance.
\begin{table}[t]
    \centering
    \small
    \resizebox{\columnwidth}{!}{
        \begin{tabular}{lcccccc}
            \toprule
            \textbf{Task}   & \multicolumn{2}{c}{\textbf{Gemma}} & \multicolumn{2}{c}{\textbf{Llama}} & \multicolumn{2}{c}{\textbf{Qwen}}                                              \\
            \cmidrule(lr){2-3} \cmidrule(lr){4-5} \cmidrule(lr){6-7}
                            & \textbf{TPR}                       & \textbf{TNR}                       & \textbf{TPR}                      & \textbf{TNR} & \textbf{TPR} & \textbf{TNR} \\
            \midrule
            Character Count & 0.99                               & 1.00                               & 0.99                              & 1.00         & 0.99         & 1.00         \\
            Term Exclusion  & 1.00                               & 0.99                               & 1.00                              & 1.00         & 1.00         & 0.99         \\
            JSON Format     & 1.00                               & 0.99                               & 1.00                              & 1.00         & 1.00         & 0.99         \\
            Word Count      & 0.95                               & 0.95                               & 0.98                              & 0.97         & 0.92         & 0.94         \\
            Register        & 0.99                               & 0.99                               & 0.99                              & 0.99         & 1.00         & 1.00         \\
            Term Inclusion  & 0.99                               & 0.98                               & 0.99                              & 0.98         & 0.94         & 0.96         \\
            Toxicity        & 0.97                               & 0.97                               & 0.98                              & 0.98         & 0.89         & 0.87         \\
            Sentiment       & 0.98                               & 0.97                               & 0.98                              & 0.99         & 0.97         & 0.94         \\
            Topic           & 0.93                               & 0.84                               & 0.92                              & 0.89         & 0.82         & 0.78         \\
            \bottomrule
        \end{tabular}
    }
    \caption{Per-class recall for the best probe of each task and model. \textbf{TPR}: recall for the compliant class (label~1). \textbf{TNR}: recall for the non-compliant class (label~0). Values near 1.0 on both columns indicate balanced probes with no systematic bias toward false positives or false negatives.}
    \label{tab:confusion_matrices}
\end{table}

\section{NormDrop robustness: random-subspace baseline}
\label{sec:app-normdrop}

A potential confound in the cross-task causal ablation (\cref{subsec:universality}, Figure~\ref{fig:norm_drop}) is that removing \emph{any} set of directions from the activation space—not only task-relevant ones—could degrade probe accuracy simply because fewer dimensions remain available to the classifier. Under this account, the Normalized Accuracy Drop (NormDrop, Equation~\ref{eq:norm_drop}) would be an artifact of the dimensionality reduction induced by nullspace projection rather than evidence of shared representational structure between tasks.

To rule out this confound, we compare against a random-subspace baseline of matched rank. For each source task~$j$, we replace its INLP nullspace projection $P_{\text{null}}^j$ with a random orthonormal projection of identical rank: given the rank~$r_j$ of the real nullspace, we sample a matrix $R \in \mathbb{R}^{d \times r_j}$ with i.i.d.\ Gaussian entries, take the orthonormal basis $Q$ from its QR decomposition, and form the random nullspace projection $P_{\text{null}}^{\text{rand}} = I - QQ^\top$. We then recompute NormDrop for all task pairs, keeping the data, the target probe, and the normalization identical to the main analysis. Crucially, this control isolates \emph{what} is removed from \emph{how much} is removed: because the random subspace has the same rank as the real nullspace, any difference between the two conditions cannot be attributed to dimensionality loss alone. If NormDrop merely reflected the projection operation, the two conditions would yield comparable, similarly distributed drops; a divergence in their structure would instead confirm that the real drops are specific to the task-relevant directions extracted by INLP.

\begin{figure*}[t]
    \centering
    \begin{subfigure}{\linewidth}
        \centering
        \includegraphics[width=\linewidth]{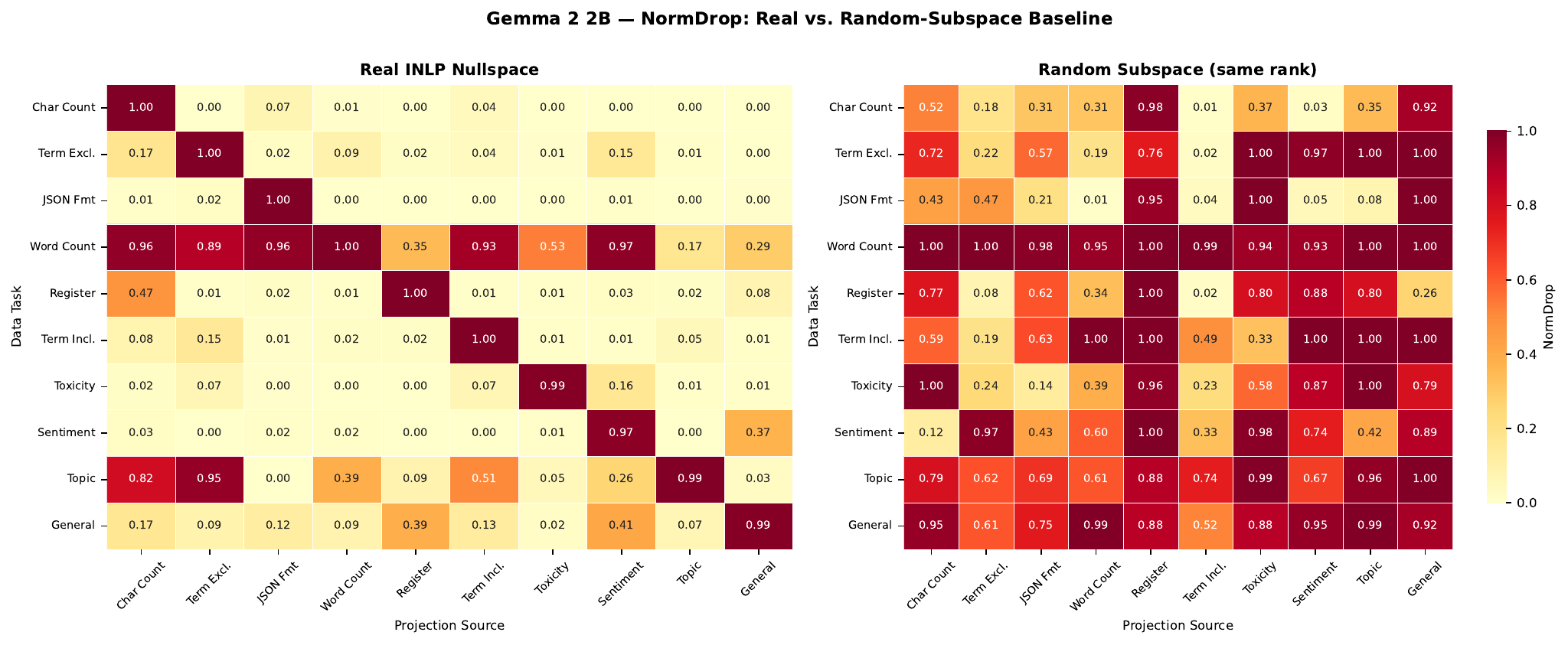}
        \caption{Gemma}
        \label{fig:normdrop_baseline_gemma}
    \end{subfigure}

    \begin{subfigure}{\linewidth}
        \centering
        \includegraphics[width=\linewidth]{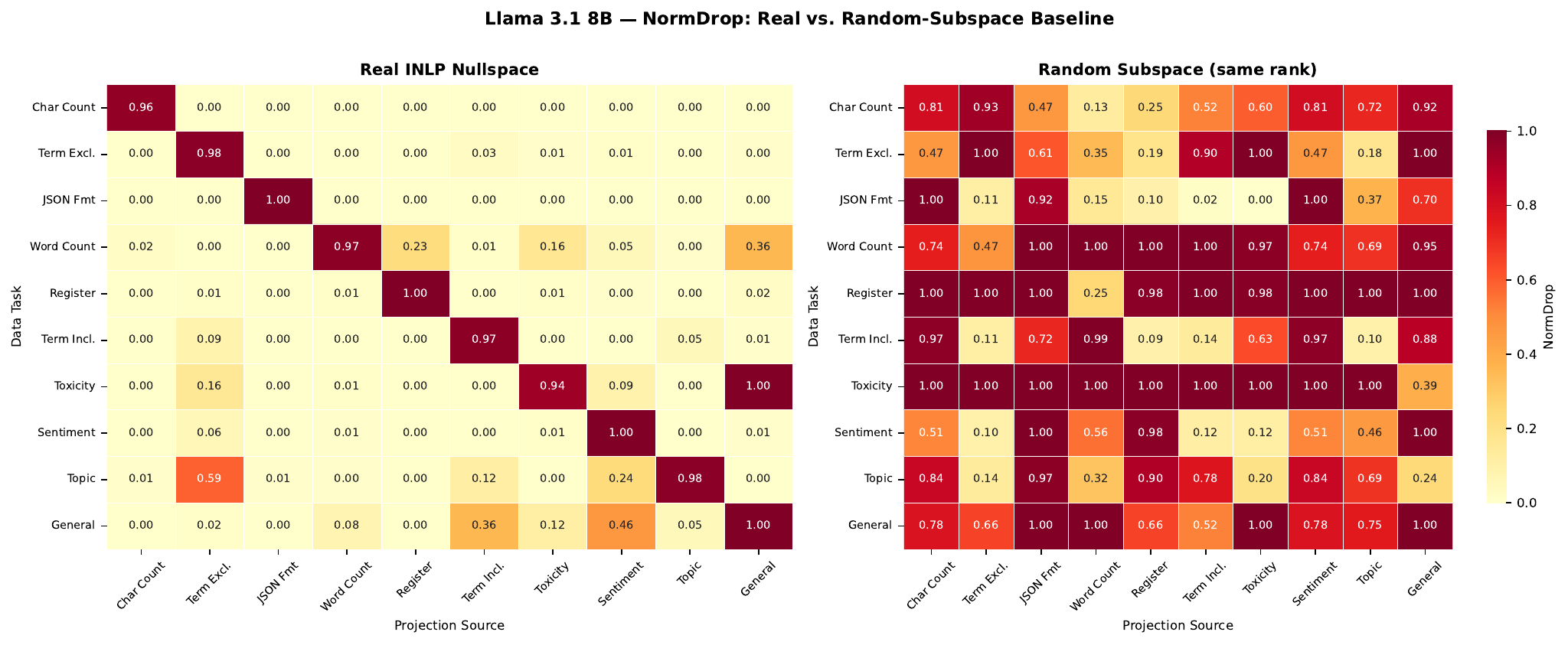}
        \caption{Llama}
        \label{fig:normdrop_baseline_llama}
    \end{subfigure}

    \begin{subfigure}{\linewidth}
        \centering
        \includegraphics[width=\linewidth]{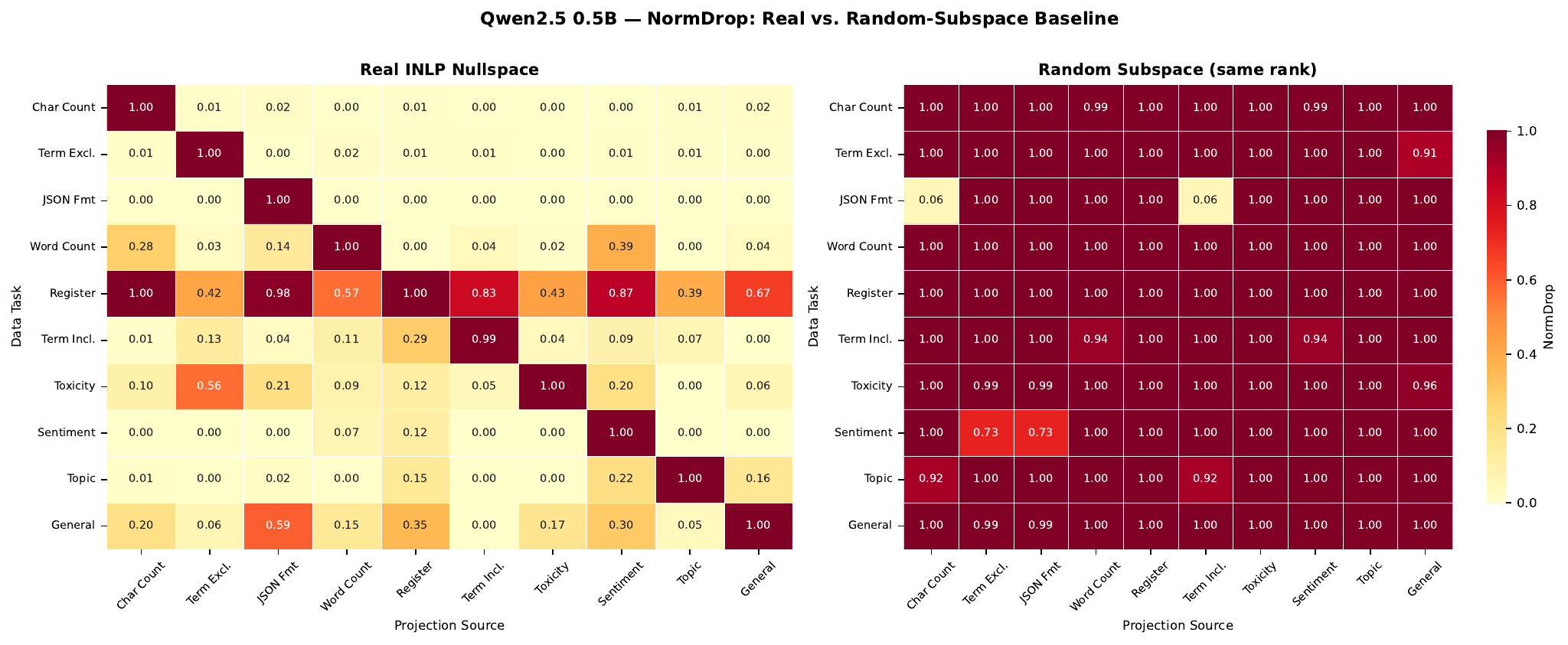}
        \caption{Qwen}
        \label{fig:normdrop_baseline_qwen}
    \end{subfigure}

    \caption{Normalized accuracy drop (NormDrop) for the real INLP nullspace ablation vs.\ random-projection baseline across all three models (EOS scope). Each cell $(i,j)$ shows the NormDrop of probe~$i$ after removing probe~$j$'s information via nullspace projection or random projection. The real ablation (left panels) is sparse and diagonal-dominated; the rank-matched random baseline (right panels) produces diffuse, non-specific drops, confirming that the real dependencies are specific to the task-relevant directions extracted by INLP rather than an artifact of removing an equal number of arbitrary directions.}
    \label{fig:normdrop_baseline}
\end{figure*}

Figure~\ref{fig:normdrop_baseline} reports the real NormDrop matrices alongside the random-subspace baseline for all three models at the EOS scope. The two conditions produce categorically different signatures. The real INLP ablation is sparse and diagonal-dominated: removing a task's own subspace collapses its probe to chance (diagonal $\approx 1.00$), while off-diagonal drops are mostly low and concentrated on a small number of interpretable cross-task dependencies. The random baseline, in contrast, produces diffuse, non-specific degradation that is elevated across the matrix. This dissociation is the relevant evidence: a generic effect of removing $r_j$ arbitrary directions would degrade probes indiscriminately (as the random baseline does), whereas the real procedure degrades a target probe only when the removed subspace carries information that probe relies on. The contrast is sharpest for Qwen and Llama, where random removal is broadly destructive but real removal is highly localized; for Gemma the dissociation is weaker but the same qualitative pattern holds. We note that Word Count is sensitive to subspace removal under both conditions across models, indicating that this particular probe is more fragile to dimensionality loss in general; the conclusion therefore rests on the aggregate sparsity of the off-diagonal structure rather than on every individual cell.

\section{Disentangling the instruction from the chat template}
\label{sec:app-template}

The specificity analysis of \cref{subsec:specificity} contrasts a constrained task against a null baseline of open-ended generation. These two conditions differ along two axes at once: whether an instruction is present, and whether the model's chat template is applied. This appendix separates the two.

We consider the full $2\times2$ design in Table~\ref{tab:template_design}. Conditions~A and~D are the two reported in the main text; conditions~B and~C are added here. Neither B nor C is a normal operating condition for an instruction-tuned model---a model is not usually given an instruction stripped of its template, nor an open-ended continuation wrapped in one---so we treat the results below as diagnostic estimates of how strong the formatting effect is, rather than as clean measurements of a well-defined state.

\begin{table}[h]
    \centering
    \small
    \begin{tabular}{@{}lll@{}}
        \toprule
                             & \textbf{With template}   & \textbf{Without template}      \\
        \midrule
        \textbf{Instruction} & A: constrained task      & B: constrained task,           \\
                             & \textit{(reported)}      & raw text \textit{(new)}        \\
        \addlinespace
        \textbf{No instr.}   & C: continuation in       & D: open-ended                  \\
                             & user turn \textit{(new)} & generation \textit{(reported)} \\
        \bottomrule
    \end{tabular}
    \caption{The $2\times2$ design separating the instruction from the chat template. A and D are the conditions reported in \cref{subsec:specificity}; B and C are added here.}
    \label{tab:template_design}
\end{table}

Content is held fixed across cells so that only the axis of interest varies: A and~C share the same continuation strings, and A and~B share the same instruction and response strings. The ablation requires no retraining. Intensity (Equation~\ref{eq:intensity}) is computed against the rowspace $P_{\text{row}}$ already extracted in the main analysis, so the new conditions supply activations only. Every cell is read at the \textbf{EOS token}, the position on which all rowspaces were fitted.

\paragraph{Decomposition.} The comparison reported in the main text, $\text{A}-\text{D}$, changes both axes simultaneously. Adding condition~C splits it into the two effects separately:
\begin{itemize}
    \item $\text{A}-\text{C}$ isolates the \textbf{instruction}, holding the template fixed;
    \item $\text{C}-\text{D}$ isolates the \textbf{template}, holding the (open-ended) content fixed.
\end{itemize}
Both are computed the same way: per task, as the difference in normalized intensity ($\|X P_{\text{row}}^\top\| / \|X\|$) between the two conditions, then summarized as the median across the nine tasks. A positive value means the first condition raises the constraint signal above the second; a value near zero means no detectable effect along that axis.

\begin{table}[h]
    \centering
    \small
    \begin{tabular}{@{}lcc@{}}
        \toprule
        \textbf{Model} & \textbf{Instruction} ($\text{A}-\text{C}$) & \textbf{Template} ($\text{C}-\text{D}$) \\
        \midrule
        Gemma          & $+0.063$ \quad (7/9)                       & $+0.096$ \quad (5/9)                    \\
        Llama          & $+0.053$ \quad (8/9)                       & $+0.197$ \quad (8/9)                    \\
        Qwen           & $-0.011$ \quad (1/9)                       & $-0.032$ \quad (2/9)                    \\
        \bottomrule
    \end{tabular}
    \caption{The $\text{A}-\text{C}$ / $\text{C}-\text{D}$ decomposition: instruction effect with the template held fixed, and template effect with (open-ended) content held fixed. Each cell is the median across the nine tasks of the per-task normalized-intensity difference, with the number of tasks showing a positive effect in parentheses.}
    \label{tab:template_ablation}
\end{table}

\paragraph{The instruction effect survives the control.} Removing the template from the open-ended condition does affect how the input is processed. Crucially, however, it does not account for the whole reported gap: with the template held fixed, removing only the instruction (A~$\rightarrow$~C) still lowers intensity for Gemma and Llama, consistently across 7 of 9 and 8 of 9 tasks respectively (Table~\ref{tab:template_ablation}). A genuine instruction component therefore remains beyond the template, and the constrained-versus-null gap reported in \cref{subsec:specificity} is not a mere formatting artifact.

\paragraph{The model-dependent pattern is preserved.} Ranked by per-task consistency the ordering is Llama (8/9) $\geq$ Gemma (7/9) $\gg$ Qwen (1/9). Qwen remains the exception whose null condition meets or exceeds the constrained one, exactly as reported in \cref{subsec:specificity}. The template does contribute to intensity in its own right, which is precisely the entanglement with general language-modeling features that \cref{subsec:specificity} describes, rather than a counter-result.

\paragraph{The template effect is often the larger of the two.} Table~\ref{tab:template_ablation} makes the relative size of the two axes explicit, and it is worth stating plainly: for Gemma and Llama, $\text{C}-\text{D}$ exceeds $\text{A}-\text{C}$ in magnitude---by roughly $1.5\times$ for Gemma and $3.7\times$ for Llama---and for Qwen the (negative) template effect is likewise larger than the instruction effect. Formatting is not a minor nuisance term alongside a dominant instruction signal; on this evidence it is at least as large a driver of intensity as the instruction itself, for every model tested. This does not undercut the previous paragraph's conclusion that an instruction-specific component remains, but it does mean that component is the smaller of the two contributions we can measure, and should be described that way rather than as the primary source of the reported gap.

Median magnitude and directional consistency come apart, however, and for one model they point in opposite directions. Gemma's template effect has the larger median but is positive in only 5 of 9 tasks, against 7 of 9 for its instruction effect: the larger of the two axes is also the less reliable one in sign. Llama is the model for which the template dominates on both measures (8/9, median $+0.197$). We therefore read the magnitudes as evidence about how much of the reported gap formatting can account for, and the counts as evidence about which axis acts consistently across tasks; only for Llama do both favour the template.

\paragraph{The contrast collapses without the template.} Consistent with condition~B not being a well-defined operating condition, the instruction effect measured without the template is approximately zero: $\text{B}-\text{D}$ is $+0.003$, $+0.000$ and $-0.003$ for Gemma, Llama and Qwen respectively. Because $P_{\text{row}}$ is fitted on templated activations it is a template-specific direction, with little for raw text to project onto. Condition~B is reported for completeness, but its near-zero contrast is ambiguous between ``no instruction effect'' and ``the probe direction does not transfer to raw text'', and we do not read it as evidence in either direction. The in-distribution cell~C carries no such ambiguity and is the one that bears on the confound.

\paragraph{Remark.} A residual confound, an effect partly carried by some correlated signal, is intrinsic to activation-level mechanistic analysis and is rarely eliminable in full. The decomposition above addresses the most plausible such confound in our setting: it shows that the template accounts for a substantial share of the originally reported $\text{A}-\text{D}$ gap, which is precisely why the two axes need to be separated, and it isolates the residual instruction component that survives once they are.

\end{document}